\documentclass[a4paper, 
10pt
]{article}

\usepackage[a4paper,left=3cm,right=3cm,top=2.5cm,bottom=2.5cm]{geometry}
\usepackage{amsmath}
\usepackage{multirow}
\usepackage[table]{xcolor}
\usepackage{amsopn}
\usepackage{mathtools}
\usepackage{comment}
\usepackage{booktabs}
\usepackage{algorithm}
\usepackage{algpseudocode}
\usepackage{amssymb}
\usepackage{url}
\usepackage{bm}
\usepackage{tabularx}
\usepackage{enumerate}
\usepackage{enumitem}
\usepackage{array}
\usepackage{tikz}
\usepackage[export]{adjustbox}
\usepackage{tikzsymbols}
\usepackage{authblk}
\numberwithin{equation}{subsection}
\usepackage{xcolor}
\usepackage[hidelinks]{hyperref}
\usepackage{cleveref}
\usepackage{chngcntr}
\usepackage{xfrac}

\definecolor{mediumpurple}{rgb}{0.58, 0.0, 0.83}
\numberwithin{equation}{subsection}
\renewcommand{\nameref}[1]{``\oldnameref{#1}''}
\newcommand{\reviewerA}[1]{#1}
\newcommand{\reviewerB}[1]{#1}
\newcommand{\reviewerALL}[1]{#1}

\renewcommand{\nameref}[1]{\ref{#1}}

\begin{document}

\title{Generative Adversarial Reduced Order Modelling}
\author[1]{Dario Coscia\footnote{dario.coscia@sissa.it}}
\author[1]{Nicola~Demo\footnote{nicola.demo@sissa.it}}
\author[1]{Gianluigi~Rozza\footnote{gianluigi.rozza@sissa.it}}

\affil[1]{Mathematics Area, mathLab, SISSA, via Bonomea 265, I-34136, Trieste, Italy}
\maketitle

\begin{abstract}
   In this work, we present GAROM, a new approach for reduced order modeling (ROM) based on generative adversarial networks (GANs). GANs \reviewerB{attempt to learn to generate data with the same statistics of the underlying distribution of a dataset, using two neural networks, namely discriminator and generator}. While widely applied in many areas of deep learning, little research is done on their application for ROM, i.e. approximating a high-fidelity model with a simpler one. In this work, we combine the GAN and ROM framework, introducing a data-driven generative adversarial model able to learn solutions to parametric differential equations. \reviewerB{In the presented methodology, the discriminator is modeled as an autoencoder, extracting relevant features of the input, and a conditioning mechanism is applied to the generator and discriminator networks specifying the differential equation parameters}. We show how to apply our methodology for inference, provide experimental evidence of the model generalization, and perform a convergence study of the method.

\end{abstract}

\section{Introduction}
Partial differential equations (PDEs) can model complex physical processes, ranging from fluid dynamics, and aerospace engineering to material science \cite{strauss2007partial, bauer2015quiet, lomax2001fundamentals}. Nevertheless, only specific types of PDEs can be analytically solved, while most require numerical approximations such as finite  difference or finite element methods. \reviewerB{Numerical solvers, as the aforementioned ones, have many qualities such as error estimation for the solution, stability for long-time simulations, and generalization across different boundary/ initial conditions or geometries \cite{morton2005numerical, butcher2016numerical}. Nevertheless, these numerical approaches are extremely costly regarding computational resources, requiring significant time to obtain simulation results.} This is even worse in parametric contexts, where a new simulation must be performed for each set of new parameters, making unfeasible real-time computations. To overcome this barrier, reduced order models (ROMs) have become an emerging field in computational sciences, fulfilling the demand for efficient computational tools for performing real-time simulations~\cite{hesthaven2016certified, lassila2014model, lucia2004reduced, kadeethum2022enhancing, joshi2019generative}. \reviewerB{In particular, data-driven ROMs have grasped increasing attention from the scientific community due to their ability to generalize across multiple system configurations using only a finite set of data. This modeling approach is ``agnostic'' in the sense that the equations describing the phenomenon are not required, rendering the methodology versatile even when data originate from real-world observations}. 

\reviewerB{ROMs for parametric PDEs are usually composed of two steps: dimensionality reduction, and manifold interpolation. In the dimensionality reduction phase, a low-dimensional space is employed to optimally represent the problem at hand; while the manifold interpolation procedure is used for mapping the parameter space to the mode of variation coefficients space. One of the most used techniques for the ROM's (linear) dimensionality reduction phase is the proper orthogonal decomposition (POD)\cite{berkooz1993proper}, while radial basis functions (RBFs) \cite{lazzaro2002radial} are usually used as a standard interpolation procedure. Data-driven ROM using linear dimensionality reduction like POD has the major drawback that exhibits limitations for non-linear dynamics \cite{csala2022comparing}}.

Deep learning (DL) algorithms have recently emerged as new approaches for performing non-linear ROM using autoencoders~\cite{kadeethum2022reduced, maulik2021reduced, eivazi2020deep, demo2023deeponet}. These architectures consist of two neural networks, namely encoder and decoder. The encoder network is used for compressing the data onto a lower-dimensional manifold, while the decoder performs the opposite transformation. The final objective is to minimize the reconstruction error between the encoded-decoded data and the original one, quantified by a specific norm. \reviewerB{For interpolating the reduced manifold, the RBF procedure can be used, or neural networks, such as long short-term memory (LSTM) networks for temporal predictions \cite{mohan2018deep, gonnella2023two}, or feed-forward neural networks in parametric problems \cite{hesthaven2018non} can be employed}. 

Regardless of the great advancement achieved in generalizability by neural networks \cite{romor2022non, fresca2021comprehensive, gruber2022comparison, lu2021learning, seidman2022nomad}, only discriminative models have been extensively investigated. Discriminative models, such as regression techniques, are machine learning models optimized for \reviewerB{learning the conditional probability of obtaining an output given a specific input. In other words, they focus on learning decision boundaries that separate (discriminative part) the data during training. During inference, the new data point is mapped to the respective boundary, depending on a specific distance metric \cite{goodfellow2016deep}}. Nevertheless, these models lack a semantic understanding of the phenomenology characterizing the data \cite{tomczak2022deep}, and quantifying the uncertainty may be challenging.
On the other hand, \reviewerB{generative modeling is used to model the dataset probability distribution using a probabilistic approach. Indeed, generative models aim to learn how to generate new data with the same statistics as the underlying dataset distribution}, which forces the network to learn specific patterns in the data. Latent variable models~\cite{bishop1998latent} are an example of generative models that have been successfully applied in many fields, see \cite{rombach2022high, ilse2020diva, oord2016wavenet, razavi2019generating} as far from an exhaustive list of possible applications. 

Recently, latent variable model approaches, specifically variational autoencoder (VAE) models \cite{kingma2013auto}, have been applied for reduced order modeling. In particular, \reviewerB{in \cite{solera2023beta} a ROM for fluid flow predictions is presented. The model uses a VAE architecture for dimensionality reduction to produce near-orthogonal representations as in \cite{eivazi2022towards}, while the interpolation is done using a transformer networks\cite{vaswani2017attention}}. VAEs represent a powerful subclass of latent variable models, namely prescribed models. The (parametric) distributions defining the probabilistic model must be chosen upfront in prescribed models. 
Nevertheless, the quality of the model can be affected if too simplistic distributions are used \cite{tomczak2022deep, rezende2018taming}. 

The implicit models --- 
different from the above-mentioned ones
--- 
distribution. In particular in GANs, the objective is achieved by implicitly modelling the true unknown probability distribution through a generator network trained by adversarial learning. 

\reviewerALL{The usage of GANs for solving PDEs is an emerging research line, and several works have already addressed the problem. Specifically, in\cite{malik2020learning} a GAN is used to learn solutions across multiple initial conditions for a specific PDE, by combining the GAN loss with a physics-informed \cite{raissi2019physics} one. Other examples of physics-informed GAN are the works in\cite{daw2021pid, randle2020unsupervised, yang2019highly}. The first one\cite{daw2021pid}, trained by minimizing the classical GAN loss, injects physical information during training by augmenting the generator and discriminator with a physics consistency score, assessing how physical the produced generator field is. The second one\cite{randle2020unsupervised}, on the other hand, uses the GAN architecture to learn the loss for a physics-informed neural network \cite{raissi2019physics}, i.e. the training is unsupervised. Finally, the third one\cite{yang2019highly} uses a variation of the methodology in\cite{daw2021pid} to learn stochastic PDEs. \reviewerA{In contrast} from the models above where the methodology is limited to learning solutions for a fixed PDE, the works in \cite{kadeethum2021framework, kadeethum2022continuous} introduce a way to condition the generative process for parametric and time-dependent PDEs. A common point of the mentioned methodologies is the fact that vanilla GANs, as introduced in\cite{goodfellow2020generative}, are used in training with the addition of physical information. However, in \cite{kemna2023reduced} the authors show that vanilla GANs, i.e. without any physical information, obtain in general higher errors compared to discriminative models. Since the addition of physical loss to the GAN loss can always be done, we seek in the article to introduce a methodology for solving parametric PDEs that can compete against discriminative models by leveraging only the GAN loss. In addition, differently from\cite{kadeethum2021framework, kadeethum2022continuous} we want the generator to not be deterministic, i.e. we want to introduce stochasticity in the network input in order to perform uncertainty quantification as done in\cite{silva2023generative, he2023survey}.}

In this work, we present GAROM, a generative adversarial reduce order model. GAROM is a novel reduced order model framework, based on a variation of conditional boundary equilibrium GAN \cite{conditional_began}. In particular, GAROM uses adversarial learning to learn the distribution over high-fidelity data, conditioned on domain-specific conditioning. At convergence, the GAROM network is able to generate high-fidelity data given domain-specific conditioning, e.g. PDE parameters, or temporal time steps. We also present a regularized version of the methodology, r-GAROM, providing extra information on the generative process as explained in section~\nameref{sec:methods}. The frameworks are tested in a variety of problems, comparing the prediction ability to state-of-the-art ROM methods, and analyzing the convergence robustness. Empirical results on the network's ability to generalize to unseen data are provided, and different uncertainty quantification strategies based on statistics are presented. Finally, we provide further possible research directions to investigate this novel methodology.

\section{Methods}\label{sec:methods}
\subsection{Conditional Boundary Equilibrium GAN}
Boundary equilibrium generative adversarial network (BEGAN) \cite{berthelot2017began} is an autoencoder based generative adversarial network. The BEGAN model consists of a generator network $G : \mathbb{R}^{N_z} \rightarrow \mathbb{R}^{N_u}$, generating a domain-specific sample in $\mathbb{R}^{N_u}$ by passing random noise $\mathbf{z}\in\mathbb{R}^{N_z} $ following $p(\mathbf{z})$ distribution; and a discriminator network $D: \mathbb{R}^{N_u} \rightarrow \mathbb{R}^{N_u}$, which \reviewerB{encodes and decodes} real and generated samples. The core idea of BEGAN is utilizing the autoencoder reconstruction loss distribution derived from the Wasserstein distance \cite{arjovsky2017wasserstein} to approximate the data distribution. Specifically, the optimization is done with respect to the Wasserstein distance between the autoencoder reconstruction losses of real and generated samples. In addition, to prevent the imbalance of $G$ and $D$, an equilibrium term is added in the objective function.

Formally, \reviewerB{let the autoencoder reconstruction loss} $\mathcal{L}: \mathbb{R}^{N_u} \rightarrow \mathbb{R}^{+}$ defined as:
\begin{equation}
    \mathcal{L}(\mathbf{u}) = | \mathbf{u} - D(\mathbf{u}) | \quad \text{where } \mathbf{u} \in \mathbb{R}^{N_u},
\end{equation}
with $\mathbb{R}^{N_u}$ the space of real data samples. Thus, the BEGAN objective is:
\begin{equation}\label{eqn:began_objective}
    \begin{cases}
        \mathcal{L}_D = \mathcal{L}(\mathbf{u}) - k_t\, \mathcal{L}(G(\mathbf{z})) &\quad \text{minimize for }\theta_D \\
        \mathcal{L}_G = \mathcal{L}(G(\mathbf{z})) &\quad \text{minimize for }\theta_G \\
        k_{t+1} = k_t + \lambda\, (\gamma \mathcal{L}(\mathbf{u}) - \mathcal{L}(G(\mathbf{z}))) &\quad \text{for each training step $t$},
    \end{cases}
\end{equation}
with $\theta_G$ and $\theta_D$ the network parameters for the generator and discriminator, $k_t$ a control term allowing the losses balance at each step $t$, and $\lambda$ the learning rate for $k_t$. Finally, $\gamma \in [0, 1]$ is the diversity ratio defined as the ratio between the generated and real data reconstruction loss expected value:
\begin{equation}
    \gamma = \frac{\mathbb{E}\left[\mathcal{L}(G(\mathbf{z}))\right]}{\mathbb{E}\left[\mathcal{L}(\mathbf{u})\right]}.
\end{equation}
Therefore, the discriminator has two main goals: \reviewerB{encoding and decoding} real data, and discriminating real from generated data. The $\gamma$ ratio is used to balance the two objectives. In fact, for lower values of $\gamma$, the discriminator focuses more heavily on autoencoding real data, leading to smaller data diversity. BEGAN has been proved to be effective for generating high resolution images, but it lacks to specifically condition the network.

Conditional BEGAN (cBEGAN) \cite{conditional_began} has been introduced to overcome this issue. The cBEGAN objective is very similar to the one presented in Equation \eqref{eqn:began_objective}, but the generator and discriminator are conditioned on conditioning variables $\mathbf{c}\in\mathbb{R}^{N_c}$. Hence, the cBEGAN objective becomes:
\begin{equation}\label{eqn:cbegan_objective}
    \begin{cases}
        \mathcal{L}_D = \mathcal{L}(\mathbf{u} \mid \mathbf{c}) - k_t\, \mathcal{L}(G(\mathbf{z} \mid \mathbf{c})) &\quad \text{minimize for }\theta_D, \\
        \mathcal{L}_G = \mathcal{L}(G(\mathbf{z} \mid \mathbf{c})) &\quad \text{minimize for }\theta_G, \\
        k_{t+1} = k_t + \lambda\, (\gamma \mathcal{L}(\mathbf{u} \mid \mathbf{c}) - \mathcal{L}(G(\mathbf{z}\mid \mathbf{c}))) &\quad \text{for each training step $t$},
    \end{cases}
\end{equation}
with the autoencoder reconstruction loss:
\begin{equation}
    \mathcal{L}(\mathbf{u} \mid \mathbf{c}) = | \mathbf{u} - D(\mathbf{u} \mid \mathbf{c}) |.
\end{equation}
In practice, the generator is conditioned by concatenating the random noise vector $\mathbf{z}$ with the conditioning variable $\mathbf{c}$. The concatenation of the two represents the input for $G$. On the other hand, the discriminator is conditioned by concatenating the encoder output  with the conditioning variable, before passing the concatenation to the decoder.

\subsection{GAROM implementation details}
The final objective of GAROM is to obtain a unique neural network which can perform ROM, while maintaining high reconstruction accuracy and generalization. We follow a \emph{data-driven} approach for ROM, where a sample of high fidelity results from a domain-specific simulation is collected and used for training. Let $\mathbf{u} \in \mathbb{R}^{N_u}$ indicating the output of the real simulation, and $\mathbf{c} \in \mathbb{R}^{N_c}$ the variable affecting the simulation, e.g. parameters, time steps or boundary conditions. We name the variables $\mathbf{c}$ as the conditioning variables, represented in this work by the free simulation parameters, as explained in section~\nameref{sec:dataset}. 

GAROM framework is based on an implicit generative modeling approach, specifically using cBEGAN. Indeed, cBEGAN allows the discriminator to learn a latent space for the real data manifold by autoencoding, forcing the generator to learn fundamental patterns of the data. Furthermore, having defined the discriminator as an autoencoder allows to impose constrains on the latent space, e.g. orthogonality, or to change the architecture to make it more robust, e.g. utilizing denoising \cite{vincent2008extracting} or contractive \cite{rifai2011contractive} regularization. These adjustments would not be possible with a vanilla GAN \cite{goodfellow2020generative}, where the discriminator outputs a real number in $[0,1]$ only.

Formally, GAROM aims to approximate the density of the high fidelity solution data given the conditioning variables. In the present work the uniqueness of the high fidelity solutions is assumed, thus for a given parameter $\mathbf{c}$ a unique solution $\mathbf{u}$ is obtained. Following the cBEGAN framework, the GAROM objective is defined as:
\begin{equation}\label{eqn:garom_objective}
    \begin{cases}
        \mathcal{L}_D = \mathcal{L}(\mathbf{u} \mid \mathbf{c}) - k_t\, \mathcal{L}(G(\mathbf{z} \mid \mathbf{c})) &\quad \text{minimize for }\theta_D \\
        \mathcal{L}_G = \mathcal{L}(G(\mathbf{z} \mid \mathbf{c})) + \eta\, |\mathbf{u} - G(\mathbf{z} \mid \mathbf{c})| &\quad \text{minimize for }\theta_G \\
        k_{t+1} = k_t + \lambda\, (\gamma \mathcal{L}(\mathbf{u} \mid \mathbf{c}) - \mathcal{L}(G(\mathbf{z}\mid \mathbf{c}))) &\quad \text{for each training step $t$},
    \end{cases}
\end{equation}
where $\eta \in \{ 0,1 \}$ ensure the absence or presence of the \reviewerB{regularization parameter}. Indeed, when $\eta=0$ the GAROM objective is the same as the cBEGAN objective, which forces the generator to learn a distribution without any information of the uniqueness of the solution. \reviewerA{On the contrary}, setting $\eta=1$ forces the generator to learn a unique distribution, which results in a shrinkage of the generator's variance, as well as better accuracy and generalization (see Section \ref{uq}). We refer to this model as the regularized generative adversarial reduced order model, r-GAROM. We want to highlight that, \reviewerA{on the contrary} to many (classical or deep learning based) ROM frameworks, we do not project onto a lower dimensional space and, once the latent representation is learnt, train an interpolator. Instead, no interpolation is done since the conditioning is passed directly to the generator and discriminator. Hence, we perform only one neural network training with our methodology. 

\begin{figure}[!ht]
    \centering
\resizebox{\columnwidth}{!}{%
\input{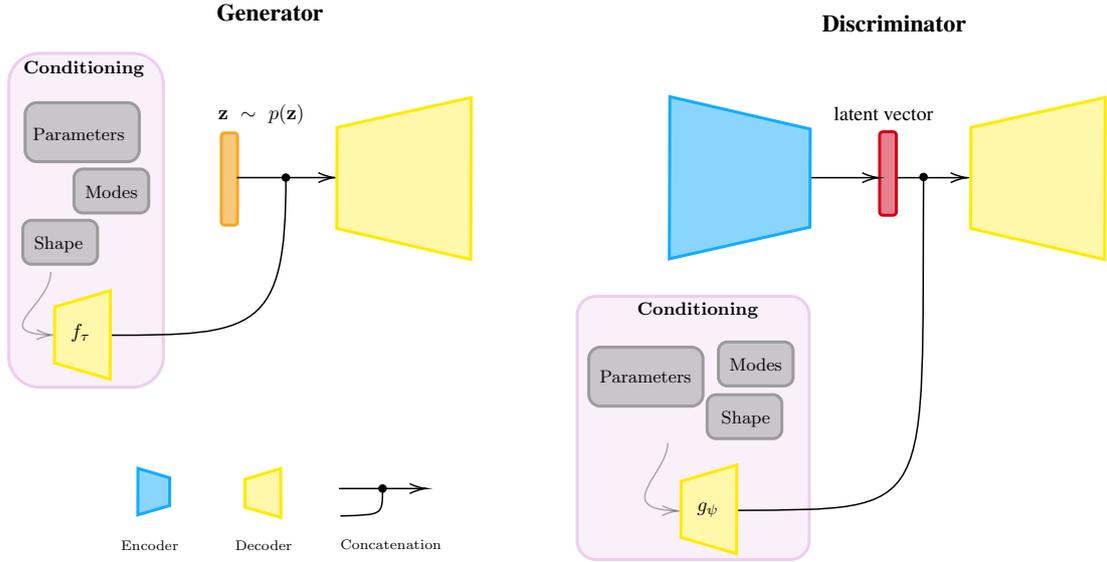}
}
\caption{\textbf{A schematic representation for GAROM generator and discriminator}. The \emph{Generator} input is the concatenation of random noise $\mathbf{z}$, and the conditioning representation $f_\tau(\mathbf{c})$. The \emph{Discriminator} encodes the input obtaining a latent vector, which is concatenated with the conditioning representation $g_\psi(\mathbf{c})$ before it is passed to the decoder. \label{fig:garom}}
\end{figure}

The (r-)GAROM architecture is depicted schematically in Figure \ref{fig:garom}. Notice that the conditioning variables (the PDE parameters) are not concatenated directly, as done in cBEGAN, but passed to a domain-specific decoder for the generator $f_\tau$, and discriminator $g_\psi$. This is done in order to pass more information to the generator similarly to \cite{reed2016generative}, and to avoid a high mismatch in the dimension of $\mathbf{z}$ and $\mathbf{c}$. Finally, it is worth noticing that in general different conditioning variables can be applied, e.g. parameters, POD modes or coordinates information. As a consequence, a specific network is needed to encode the information into one high-dimensional vector. In the present work, only parametric problems are considered, and we leave for future works the investigation of highly effective conditioning mechanisms.

\subsection{GAROM predictive distribution}
\label{sec:prediction}
Once the (r-)GAROM model is trained, a probabilistic ensemble for the output variable $\mathbf{u}(\mathbf{c})$ can be constructed leveraging $G(\mathbf{z}\mid\mathbf{c})$. Following a similar approach to \cite{yang2019adversarial}, we can predict the mean prediction $\hat{\mathbf{u}}$ and its variance $\hat{\boldsymbol{\sigma}}^2$, for a new conditioning variable $\mathbf{c}^*$, by Monte Carlo sampling:
\begin{equation}
\begin{split}
    \hat{\mathbf{u}} &= \mathbb{E}_{p_G}[\mathbf{u} \mid \mathbf{z}, \mathbf{c}^*] \approx \frac{1}{K} \sum_{k=1}^K G(\mathbf{z}_k \mid \mathbf{c}^*), \\
    \hat{\boldsymbol{\sigma}}^2 &= \mathbb{V}\text{ar}_{p_G}[\mathbf{u} \mid \mathbf{z}, \mathbf{c}^*] \approx \frac{1}{K-1} \sum_{k=1}^K [\hat{\mathbf{u}} - G(\mathbf{z}_k \mid \mathbf{c}^*)]^2.
\end{split}
\end{equation}
In particular, $p_G$ is the implicit distribution learned by the generator via adversarial optimization, $K$ is the number of Monte Carlo samples used for approximating the distribution moments, and $\mathbf{z}_k$ are the samples from the latent distribution $p(\mathbf{z})$.

\subsection{Uncertainty quantification strategies}\label{sec:uq_strat}
\reviewerB{Quantifying the uncertainty
of a prediction} is a common problem for data-driven reduced order models \cite{rozza2022advanced}. Exploiting the predictive distribution of (r-)GAROM one can obtain moment estimates by Monte-Carlo sampling, as discussed in section~\nameref{sec:prediction}. These estimates can be used during inference to compute bounds in probability of the prediction error, i.e. the difference between real and average (r-)GAROM solution. As an example, employing the Markov inequality we can find the following bound for any threshold $a>0$:
\begin{equation}\label{eqn:markov}
    \text{Prob}[(\mathbf{u}_{\rm{true}}(\mathbf{c})-\hat{\mathbf{u}}(\mathbf{c}))^2 \geq a] \leq \frac{1}{a} \left\{ \mathbb{E}_{\rm{Prob}} \left[\mathbb{E}_{p_G}[(\mathbf{u}_{\rm{true}}(\mathbf{c})-\mathbf{u}(\mathbf{c}))^2]  - \hat{\boldsymbol{\sigma}}^2(\mathbf{c}) \right] \right\},
\end{equation}
where Prob is the probability distribution for $\mathbf{c}$, e.g. random uniform. In practice, the right hand side of Equation \eqref{eqn:markov} can be estimated, with sufficient statistics, after training by Monte Carlo approximation on the training dataset, resulting in an estimate in probability for the average prediction.

A more tight bound for each parameter $\mathbf{c}$, can be obtained assuming an approximation of the distribution $p_G$, and applying the theory of confidence interval. As an example, assuming $p_G$ is well approximated using a normal distribution $\mathcal{N}(\hat{\mathbf{u}}, \hat{\boldsymbol{\sigma}}^2)$,  one could estimate the prediction uncertainty in a similar approach to the one adopted by \cite{gundersen2021semi}. While the first bound based on Markov inequality is certain, the latter one uses an approximation of the distribution. Nonetheless, as also evidenced in \cite{gundersen2021semi}, these kinds of approximations (problem-specific) can obtain very good estimates.

\subsection{Datasets description}\label{sec:dataset}
This section is dedicated to introduce the numerical benchmarks applied for testing GAROM and comparing it to the baseline models presented in section~\nameref{sec:base_model}. For notation consistency, $\mathbf{c}$ indicates the vector of free parameters while $\mathbf{u}$ the simulation solution.
\paragraph{Parametric Gaussian}
The first test case is a simple problem representing a Gaussian function moving in a domain $\Omega = [-1, 1] \times [-1, 1]$. The high fidelity function $u : \Omega \rightarrow \mathbb{R}$ is defined as:
\begin{equation}
    u(x, y) = e^{-[(x - c_1)^2 + (y - c_2)^2]},
\end{equation}
with $\mathbf{c} = [c_1, c_2]$ the centers of the Gaussian. Practically, $u$ is evaluated on $900$ points uniformly randomly chosen in $\Omega$ for a fixed $\mathbf{c}$. The evaluation results are flattened (row-major order) in a one-dimensional vector of size $900$, namely $\mathbf{u}$. The dataset is composed by $N=400$ instances $\mathbf{c}_i$ uniformly randomly chosen in $\Omega$, resulting in $N$ high fidelity solutions $\mathbf{u}_i$.
\paragraph{Graetz Problem}
The second test case deals with the Graetz-Poiseuille problem, which models forced heat convection in a channel. The problem is characterized by two parameters: $c_1$ controlling the length of the domain, and $c_2$ the Péclet number, which takes into account the heat transfer in the domain. The full domain is $\Omega(c_1) = [0, 1 + c_1]\times[0, 1]$, with $\mathbf{c}=[c_1, c_2] \in [0.1, 10] \times [0.01, 10]$. The high fidelity solution $u : \Omega(c_1) \rightarrow \mathbb{R}$ is obtained by solving:
\begin{equation}
    \begin{cases}
        -\Delta u(x, y; \mathbf{c}) + c_2 y(1-y)\frac{\partial}{\partial x} u(x, y; \mathbf{c})=0 &\quad (x, y) \in \mathring{\Omega}(c_1)\\
        u(x=0, y; \mathbf{c})=0 &\quad y \in [0, 1]\\
        u(x, y=0; \mathbf{c})=0 &\quad x \in [0, 1]\\
        u(x, y=1; \mathbf{c})=0 &\quad x \in [0, 1]\\
        u(x, y=0; \mathbf{c})=1 &\quad x \in [1, 1+c_1]\\
        u(x, y=1; \mathbf{c})=1 &\quad x \in [1, 1+c_1]\\
        \partial_{\mathbf{n}}u(x=1+c_1, y)=0 &\quad y \in [0, 1]\\
    \end{cases}
\end{equation}
where $\Delta$ indicates the Laplacian operator, and $\partial_{\mathbf{n}}$ the normal derivative. The high-fidelity solutions are numerically computed by finite elements following the work \cite{hesthaven2016certified}, using a mesh of $5160$ points. Specifically, $N=200$ instances of $\mathbf{c}_i$ are uniformly randomly chosen in $\Omega(c_1)$, resulting in $N$ high fidelity solutions $\mathbf{u}_i$ arranged as a vector (row-major order).
\paragraph{Lid Cavity}
The last test case is the famous Lid Cavity problem, modeling isothermal, incompressible flow in a two-dimensional square domain. The domain is composed of a top wall moving along the horizontal axis, while the other three walls are stationary. At low Reynolds number (Re) the flow is laminar, but it becomes turbulent when increasing the Re. A complete mathematical formulation of the problem is found in~\cite{schreiber1983driven}. The full domain is \reviewerB{$\Omega = [d, d]\times[d, d]$} with $d=0.1 m$ the domain size. The free parameter $c$ is the magnitude of the velocity of the top wall, measured in $ms^{-1}$. The Reynolds number is given by:
\begin{equation}
    \text{Re} = \frac{d c}{\nu},
\end{equation}
with the kinematic viscosity $\nu=10^{-5}m^2s^{-1}$. To compute the high fidelity solution, we \reviewerB{simulate for $5s$ using a time step of $0.0001$} keeping only the last time-step simulation. The mesh is \reviewerB{composed of} hexahedral cells of $70 \times 70 \times 1$ number of cell points in each direction. We use a cell-center finite volume scheme. The dataset is composed by $N=300$ instances of $c \in [0.01, 1]$ evenly spaced, resulting in $N$ high fidelity solutions arranged as a vector (row-major order).

\subsection{Baseline models}
\label{sec:base_model}
We compared the performances of the (r-)GAROM approach against some of the more consolidated data-driven frameworks in ROM community\cite{rozza2022advanced}. Such models rely on machine learning and/or linear algebra techniques, resulting on different approaches from the algorithmic perspectives. In such way, we aim for the most possible fair comparison, \reviewerA{distinguishing} the GAROM advantages and disadvantages with respect to the state-of-the-art in different possible operating contexts.

The baseline comparison models are:
\begin{itemize}[noitemsep,topsep=0pt]
    \item Proper Orthogonal Decomposition with Radial Basis Function interpolation (POD-RBF)\cite{salmoiraghi2018free},
    \item Proper Orthogonal Decomposition with Artificial Neural Network interpolation (POD-NN)\cite{hesthaven2018non},
    \item Autoencoders with Radial Basis Function interpolation (AE-RBF) \cite{lee2020model},
    \item Autoencoders with Artificial Neural Network interpolation (AE-NN)\cite{IvagnesDemoRozza2023JSC, hasegawa2020machine, gonzalez2018deep}.
    \item  Conditional GAN (cGAN)\cite{mirza2014conditional}
    \item  Regularized vanilla conditional GAN (r-cGAN)\cite{kemna2023reduced}
    \item Deep Operator Networks (DeepONet) \cite{lu2021learning}
    \item Deep Operator Networks with Non-Linear Manifol Decoder (NOMAD) \cite{seidman2022nomad}
\end{itemize}
\reviewerA{In a few words, POD-RBF, POD-NN, AE-RBF, and AE-NN  use proper orthogonal decomposition and an autoencoder for dimensionality reduction, while radial basis function and artificial neural network for approximating the reduced parametric solution manifold. The POD is performed by truncated singular value decomposition with a variable rank (latent dimension in the manuscript), while for radial basis function interpolation, we employed a Gaussian Kernel with length scale equal to $1$. Differently, DeepONet and NOMAD are neural operator networks, mapping directly the parameters and field coordinates to the field solution on the specified coordinates. Finally, we also present two vanilla GAN approaches: cGAN and r-cGAN,  where at the latter we add an L1 penalty loss to the generator network\cite{kemna2023reduced}}.

\reviewerB{For the neural networks-based approaches we performed hyperparameter optimization to find the architecture with lower training mean square error. The optimization was carried out with RayTune\cite{liaw2018tune}}. \reviewerA{For the specifics on network parameters and training procedures see the Supplementary Information.}

\subsection{Software details}
The PyTorch software \cite{pytorch} is used to implement the GAROM model due to its versatility and its wide used in the deep learning community. The EZyRB software~\cite{ezyrb} and PINA software~\cite{Coscia2023} are used for performing the comparison between GAROM and state-of-the-art techniques in Section \nameref{numerics}. Openfoam \cite{of} and FeniCSx \cite{fenics} are used for the numerical simulations. An implementation of GAROM is available in PINA\cite{Coscia2023}, while the datasets used are available in the Smithers package on GitHub at \href{https://github.com/mathLab/Smithers}{https://github.com/mathLab/Smithers}.

\subsection{Training setup and model architecture}\label{sec:training}
An NVIDIA Quadro RTX 4000 GPU was used to train the GAROM models, while the GAROM comparison methods are trained on Intel CPUs. In the Graetz and Lid Cavity experiments, the GAROM and baseline models are all trained on a dataset containing $80\%$ of the total number of high fidelity simulation instances \reviewerB{chosen evenly spaced}, and tested on the remaining $20\%$. \reviewerA{On the contrary}, for the Gaussian test case, $60\%$ of the high fidelity instances \reviewerB{chosen evenly spaced} are used for training, and the remaining $40\%$ for testing. This choice is due to the higher number of instances with respect to Lid and Graetz test cases. 

The GAROM models are all trained for $20000$ epochs, with Adam optimizer using a learning rate of $0.001$ for both the discriminator and generator. The GAROM training times for the benchmarks are approximate of $45$ minutes for the parametric Gaussian, $1$ hours and $30$ minutes for Graetz, and $2$ hours for the Lid Cavity, independently of the discriminator latent dimension. In order to fit all data in GPU we used a batch size of $4$ for all tests, \reviewerB{where the value was obtained by hyper-parameter optimization using RayTune\cite{liaw2018tune} choosing from $\{4, 8, 16, 24, 64\}$. The noise vector $\mathbf{z}$ are uniform random samples in $[-1, 1]$ of dimension $13$ for the Gaussian test, $8$ for the Graetz test, and $12$  for the Lid Cavity test. Also, the noise vector dimension was tuned with RayTune, choosing a random integer in the range $[8, 16]$}. Furthermore, the $\lambda$ and $\gamma$ parameters of Equation \eqref{eqn:garom_objective} are set to $0.001$ and $0.3$ respectively, as suggested in \cite{berthelot2017began}. \reviewerB{The generator, discriminator, generator conditional encoder, and discriminator conditional encoder networks are reported in Supplementary information.}

\section{Results}\label{sec:results}
In this section, the performances of GAROM and its improved variant r-GAROM are presented. To estimate the accuracy of the presented methodology, we use the mean $l_2$ relative error $\epsilon$ between the high-fidelity solution $\mathbf{u}(\mathbf{c}) \in \mathbb{R}^{N_u}$, and the generated one $\hat{\mathbf{u}}(\mathbf{c}) \in \mathbb{R}^{N_u}$, for the parameter $\mathbf{c} \in \mathbb{R}^{N_c}$. It is formally defined as:
\begin{equation}
    \epsilon = \frac{1}{N}\sum_{i=1}^{N}\frac{\|\mathbf u(\mathbf{c}_i) - \hat{\mathbf u}(\mathbf c_i)\|^2_2}{\|\mathbf u(\mathbf{c}_i)\|^2_2},
\end{equation}
with $N$ the number of testing parameters used for the metric evaluation.

All of our experiments are performed on three different test cases, representing benchmark problems in parametric contexts. \reviewerB{Aiming at increasing complexity}, we take into account a Gaussian signal moving inside a square (algebraic), the Graetz-Poiseuille problem (linear PDE), and the Lid Cavity problem (nonlinear PDE). The datasets \reviewerB{for the training are therefore composed} of the (parameter-dependent) high-fidelity solutions, which are obtained by solving such problems with the consolidated numerical solver, as exhaustively explained in section~\nameref{sec:dataset}. Moreover, for all the test cases, different latent dimensions for the discriminator network are considered. The other architectural details for the models at hand are fully described in section~\nameref{sec:training}, together with the optimizer settings.

\subsection{Model inference}\label{numerics}
\begin{figure}[!h]
    \centering
    \includegraphics[width=0.9\linewidth]{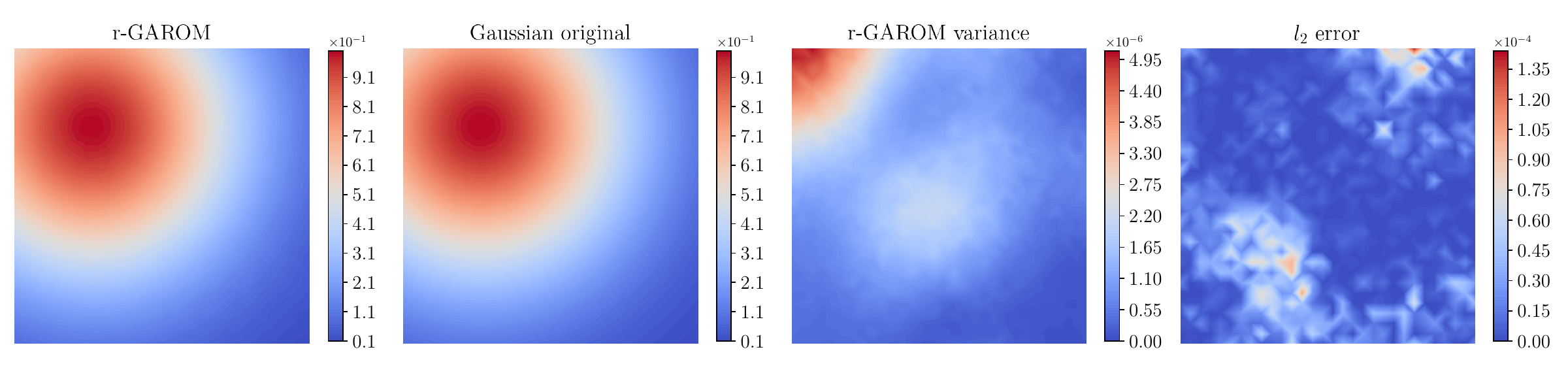}\hfill
    \includegraphics[width=0.9\linewidth]{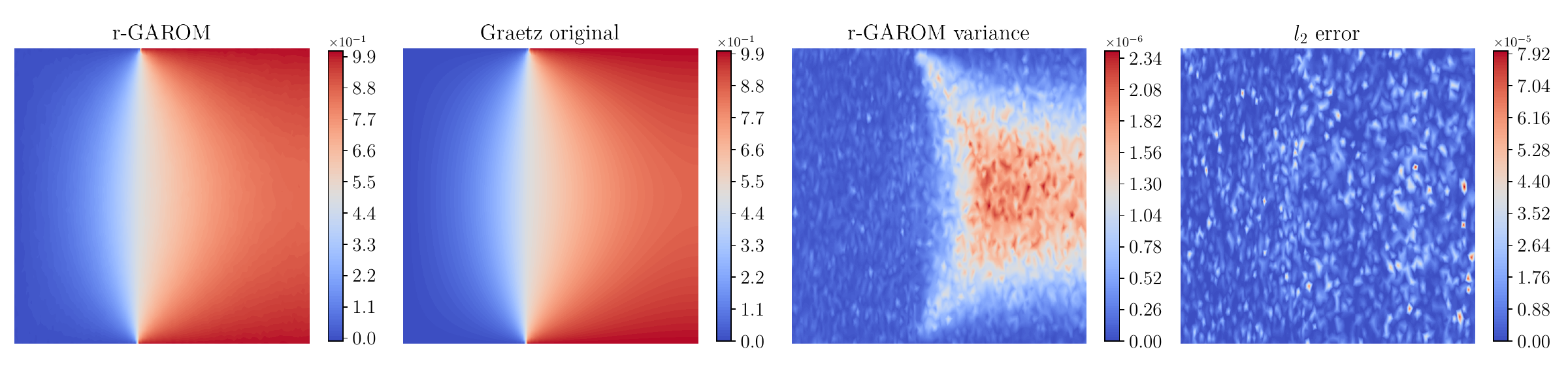}\hfill
    \includegraphics[width=0.9\linewidth]{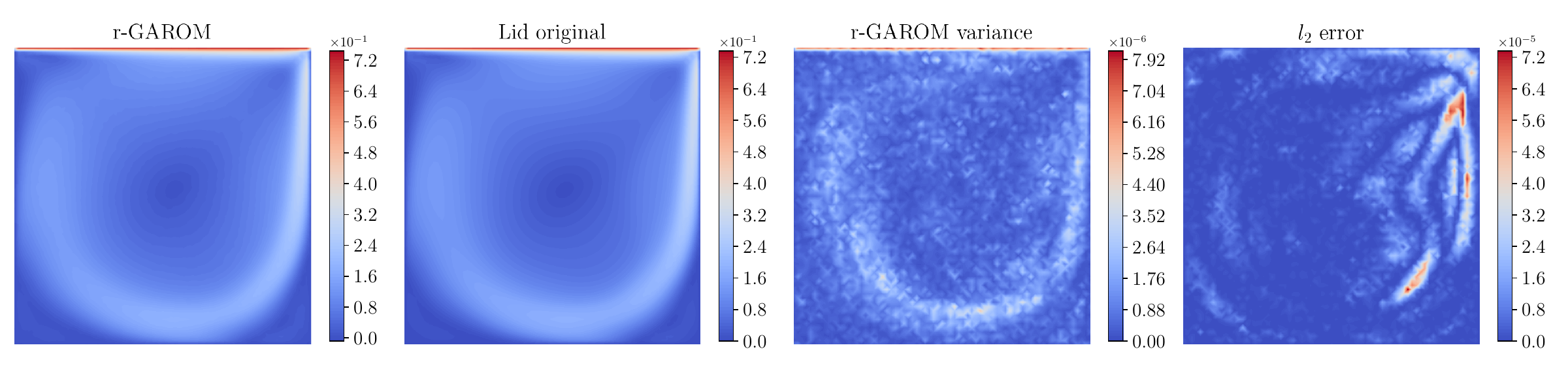}
    \caption{\textbf{r-GAROM inference results}. The images show the generated snapshot \reviewerB{representing the magnitude of the unknown field} for a testing parameter using a latent dimension of $64$, compared to the corresponding high-fidelity solution. \emph{Top}: Gaussian dataset. \emph{Center}: Graetz dataset. \emph{Bottom}: Lid cavity dataset.}
    \label{fig:uq}
\end{figure}

In the first experiment, we test the GAROM capability to predict the solution for a new parameter, aka conditioning variable.
A visualization of the r-GAROM prediction for a testing parameter, and its  $l_2$ point-wise error is reported in Figure~\ref{fig:uq}. The spatial distribution of the prediction for all the test cases is almost identical to the original one, as proved also by the low error. \reviewerB{The variance reaches low values in all the problems, demonstrating that the model has learned to properly generate the solution field. The spatial distributions of variance and error do not show similarities, but the former identifies the spatial region where the network is more uncertain (we highlight that the plots present the variance for a single test parameter), which can be potentially used to adapt the training procedure. A similar analysis like the one in Figure~\ref{fig:uq} has been replicated also for the GAROM model. The unregularized model results are barely indistinguishable to the eye with respect to the regularized counterpart, so we have not reported them. }

\reviewerA{Looking at a more precise measure for the GAROM accuracy, we calculate the mean $l_2$ relative error over a set of training and testing parameters, comparing it with the error obtained using the ROM baseline models: POD--RBF, POD--NN, AE--RBF, and AE--NN; the generative modelling baselines: cGAN, and r-cGAN; and the Operator Learning baselines: DeepONet, and NOMAD. }
As we can note from Table~\ref{tab:table}, for the majority of the experiments POD--RBF is the best across all models on the training dataset (gray rows), but it fails to properly generalize to test data.
\reviewerB{Indeed, the error computed on the test data exhibits a completely different trend with respect to the one computed over the training snapshots, especially in the Graetz case,}
showing a poor ability to correctly generalize. \reviewerA{On the contrary}, ML-based methodologies exhibit comparable train and test errors, due to their ability to learn complex nonlinear relationships by the data. For the test error, which is used to assess the model prediction ability, the r-GAROM model obtains very promising results for almost half of the test cases. Across the different deep learning models, r-GAROM obtains lower error for six out of nine tests, highlighting its encouraging precision even in this preliminary contribution.
It is also important to highlight that the (r-)GAROM architecture is fairly simple in all the numerical tests here pursued: both generator and discriminator are composed only of a few dense layers (details in section~\nameref{sec:training}). We suppose the accuracy can be further improved by using more powerful deep learning architectures, as in \cite{maulik2021reduced, dutta2022reduced, lee2020model}, at the cost of longer and more expensive training.
\begin{table}[]
    \caption{\textbf{Accuracy comparison for different test cases, methods, and latent dimensions}. The table reports the mean $l_2$ relative error $\epsilon$ (in percent). Gray rows represent the error on training data, while white rows the error on testing data. For GAROM and r-GAROM models, also the predictive standard deviation is reported (in percent). The (r-)cGAN, DeepONet and NOMAD models results do not depend on the hidden dimension, thus we report the same error value.}
    \label{tab:table} 
\vspace{0.2cm}
\resizebox{\linewidth}{!}{
\begin{tabular}{cccccccccc}
\rowcolor[HTML]{FFFFFF} 
\toprule
{\color[HTML]{000000} \textbf{Method}} & \multicolumn{3}{c}{\cellcolor[HTML]{FFFFFF}{\color[HTML]{000000} \textbf{Gaussian}}} & \multicolumn{3}{c}{\cellcolor[HTML]{FFFFFF}{\color[HTML]{000000} \textbf{Graetz}}} & \multicolumn{3}{c}{\cellcolor[HTML]{FFFFFF}\textbf{Lid Cavity}} \\ 
\cmidrule(lr){2-4} \cmidrule(lr){5-7} \cmidrule(lr){8-10}
\rowcolor[HTML]{FFFFFF} 
\multicolumn{1}{l}{\cellcolor[HTML]{FFFFFF}{\color[HTML]{000000} }} & {\color[HTML]{000000} \textbf{4}} & {\color[HTML]{000000} \textbf{16}} & {\color[HTML]{000000} \textbf{64}} & \multicolumn{1}{c}{\cellcolor[HTML]{FFFFFF}{\color[HTML]{000000} \textbf{16}}} & \multicolumn{1}{c}{\cellcolor[HTML]{FFFFFF}{\color[HTML]{000000} \textbf{64}}} & \multicolumn{1}{c}{\cellcolor[HTML]{FFFFFF}{\color[HTML]{000000} \textbf{120}}} & \multicolumn{1}{c}{\cellcolor[HTML]{FFFFFF}\textbf{16}} & \multicolumn{1}{c}{\cellcolor[HTML]{FFFFFF}\textbf{64}} & \multicolumn{1}{c}{\cellcolor[HTML]{FFFFFF}\textbf{120}}  \\
\midrule
GAROM & $(1.05\pm0.17) $ & $(0.77 \pm 0.20)  $ & $(0.88 \pm 0.16)  $ & $(0.58 \pm 0.16)  $ & $(0.55 \pm 0.14)  $ & $(0.51 \pm 0.1)  $  & $(10.7 \pm 0.22)  $ & $(9.50 \pm 0.15)  $ & $(9.91 \pm 0.17)  $  \\
r-GAROM & $\mathbf{(0.80 \pm 0.15) }$ & $(0.60 \pm 0.14)  $ & $(0.65 \pm 0.14)  $ & $\mathbf{(0.32 \pm 0.09) } $ & $\mathbf{(0.22 \pm 0.05) } $ & $\mathbf{(0.20 \pm 0.04) } $ & $(5.02 \pm 0.11)  $ & $(3.41 \pm 0.04)  $ & $(3.61 \pm 0.06)  $ \\
cGAN & $(7.72\pm0.30)$ & $(7.72\pm0.30)$  & $(7.72\pm0.30)$ & $(21.3\pm 1.10)$ & $(21.3\pm 1.10)$ & $(21.3\pm 1.10)$ & $(68.5 \pm 1.55)$ & $(68.5 \pm 1.55)$ & $(68.5 \pm 1.55)$ \\
r-cGAN & $(6.04\pm0.27)$ & $(6.04\pm0.27)$ & $(6.04\pm0.27)$ & $(1.65\pm0.05)$ & $(1.65\pm0.05)$ & $(1.65\pm0.05)$ & $(51.6\pm 1.11)$ & $(51.6\pm 1.11)$ & $(51.6\pm 1.11)$ \\
POD--RBF & $15.4 $ & $\mathbf{0.41 }$ & $\mathbf{0.25 }$ & $0.48 $ & $0.48 $ & $0.49 $ & $\mathbf{2.98 }$ & $\mathbf{1.45 }$ & $\mathbf{1.32 }$ \\
POD--NN & $15.4 $ & $1.30 $ & $1.12 $ & $0.41 $ & $0.47 $ & $0.44 $ & $3.02 $ & $2.73 $ & $3.20$ \\
AE--RBF & $2.06 $ & $0.89 $ & $1.00 $ & $1.08 $ & $1.42 $ & $1.60 $ & $6.96 $ & $3.01 $ & $3.90 $\\
AE--NN & $1.76 $ & $1.04 $  & $1.12 $ & $1.14 $ & $1.56 $ & $2.71  $ & $3.49 $ & $3.27 $ & $3.83 $\\
DeepONet & $36.1$ & $36.1$ & $36.1$ & $0.73$ & $0.73$ & $0.65 $ & $72.2$ &  $72.2$ &  $72.2$\\
NOMAD & $1.99 $ & $1.99 $  & $1.99 $ & $0.36$ & $0.36$ & $0.36$ & $63.4$ & $63.4$ & $63.4$\\
\hline
\rowcolor{gray!20}
GAROM & $(1.04\pm0.01) $ & $(0.73 \pm 0.01)  $ & $(0.86 \pm 0.01) $ & $(0.69 \pm 0.01)  $ & $(0.62 \pm 0.01)  $ & $(0.57 \pm 0.01)  $  & $(13.5 \pm 0.01)  $ & $(11.5 \pm 0.01)  $ & $(12.6 \pm 0.01)  $  \\
\rowcolor{gray!20}
r-GAROM & $\mathbf{(0.78 \pm 0.01)}$ & $(0.54 \pm 0.01)  $ & $(0.60 \pm 0.01)  $ & $(0.54 \pm 0.009)  $ & $(0.19 \pm 0.004) $ & $(0.18 \pm 0.004) $ & $(6.36 \pm 0.007) $ & $(3.75 \pm 0.003)  $ & $(4.04 \pm 0.004)  $ \\
\rowcolor{gray!20}
cGAN & $(7.04\pm0.30)$ & $(7.04\pm0.30)$  & $(7.04\pm0.30)$ & $(21.6\pm 1.10)$ & $(21.6\pm 1.10)$ & $(21.6\pm 1.10)$ & $(82.75 \pm 1.50)$ & $(82.75 \pm 1.50)$ & $(82.75 \pm 1.50)$ \\
\rowcolor{gray!20}
r-cGAN & $(5.93\pm0.27)$ & $(5.93\pm0.27)$ & $(5.93\pm0.27)$ & $(1.78\pm0.05)$ & $(1.78\pm0.05)$ & $(1.78\pm0.05)$ & $(64.90\pm 1.11)$ & $(64.90\pm 1.11)$ & $(64.90\pm 1.11)$ \\
\rowcolor{gray!20}
POD--RBF & $15.2 $ & $ \mathbf{0.23 }$ & $\mathbf{2.85\cdot10^{-5}}$ & $\mathbf{7.73 \cdot 10^{-4} }$ & $\mathbf{1.51 \cdot 10^{-11}} $ & $\mathbf{2.80 \cdot 10^{-12}} $ & $\mathbf{2.37}$ & $ \mathbf{0.29 }$ & $\mathbf{0.04}$ \\
\rowcolor{gray!20}
POD--NN & $15.2 $ & $0.65 $ & $0.59 $ & $0.36 $ & $0.38 $ & $0.40$ & $3.08 $ & $2.90 $ & $3.26 $ \\
\rowcolor{gray!20}
AE--RBF & $1.90 $ & $0.70 $ & $0.83 $ & $1.01 $ & $1.41 $ & $1.54 $ & $7.24 $ & $2.73 $ & $3.68$\\
\rowcolor{gray!20}
AE--NN & $1.40 $ & $0.82 $  & $0.96$ & $1.16 $ & $1.67 $ & $2.78   $ & $3.60 $ & $3.34 $ & $4.02 $\\
\rowcolor{gray!20}
DeepONet & $35.80 $ & $35.80 $ & $35.80 $ & $0.65 $ & $0.65 $ & $0.65 $ & $76.85$ & $76.85$ & $76.85$\\
\rowcolor{gray!20}
NOMAD & $1.98 $ & $1.98 $  & $1.98 $ & $0.42$ & $0.42$ & $0.42$ & $78.73$ & $78.73$ & $78.73$\\
\hline
  & & & & & & & &  &  \\
  & & & & & & & & \begin{tikzpicture}[scale=.4] \draw  [fill={rgb, 255:red, 155; green, 155; blue, 155 }  ,fill opacity=0.2 ] (0,0) -- (3,0) -- (3,0.6) -- (0,0.6) -- cycle ;
   \end{tikzpicture} & \begin{tikzpicture}[scale=.4] \draw  [fill={rgb, 255: red, 255; green, 255; blue, 255 }] (0,0) -- (3,0) -- (3,0.6) -- (0,0.6) -- cycle ;
   \end{tikzpicture}  \\
  & & & & & & & & Training & Testing  \\

\end{tabular}
}
\end{table}

\reviewerA{Compared to the vanilla (r-)cGAN, the (r-)GAROM methodology obtains lower error in both training and testing, with the difference in accuracy especially evidenced in the most challenging tests, i.e. the Graetz and Lid Cavity. Furthermore, the predictive standard deviation for the (r-)GAROM model is much lower than the (r-)cGAN model, empirically showing that the GAROM model is less uncertain (see \nameref{uq} for more details)}.
Finally, the r-GAROM error seems to be almost independent of the latent dimension, since only a small variation of the errors is observed in training and testing for the various dimensions. In such a way, we can empirically prove the better capability to reduce the original dimension of the snapshots, since adding new reduced dimensions does not affect the final accuracy.

\subsection{Model variance}\label{uq}
Figure~\ref{fig:uq} reports the (r-)GAROM variance obtained by a Monte Carlo approximation with different random inputs for the generator. Contrary to the ROM baseline models, GAROM can provide also an estimate of the model variance, obtained by a probabilistic ensemble (see section~\nameref{sec:prediction}). Since the solution is unique by assumption, we expect a zero variance for the generated samples given the conditioning variables. Hence, a high variance in the GAROM estimate gives an uncertain prediction of the model. 
This is confirmed by observing the variances for different models in Table~\ref{tab:table}, and the point-wise variance in Figure \ref{fig:uq}. Both GAROM and r-GAROM models report a small variance, with training variance always one or two orders of magnitude smaller than the testing one. This indicates that GAROM models are more uncertain in predicting testing data than training ones. This is expected since the GAROM model builds a distribution on the training data. However, it is worth noticing that the $l_2$ errors between training and testing in Table \ref{tab:table} are almost always identical across different dimensions, showing good generalizability of the model, as also reported in section~\nameref{gen}. It is evident from Table \ref{tab:table} that r-GAROM has a lower variance than GAROM for testing and training. This is mainly due to the regularizer term, which forces the network to converge faster, as explained in the section~\nameref{sec:methods}. 

The variance estimate, although very important for the aforementioned reasons, does not give a quantification of the predictive uncertainty, but only statistical information of the \reviewerB{learned distribution}. Nevertheless, due to the probabilistic framework adopted, we can obtain bounds in probability for the reconstruction error by using the variance estimate. Indeed, as deeply explained in section~\nameref{sec:uq_strat}, with the variance information provided by the GAROM framework, it is possible to obtain prediction uncertainty using the Markov inequality, or the confidence interval theory. While extremely important, in this work, we just show how to obtain these näive bounds. Exploring further to obtain tighter bounds, or employ the uncertainty quantification strategies for noisy data are possible new interesting research topics, which we will explore in the future.

\subsection{Generalization and robustness}\label{gen}
One fundamental aspect, especially in the context of data-driven modelling, is the ability to correctly generalize across new instances of the conditioning domain, which of course were not inside the training dataset. To empirically prove the GAROM ability to generalize over the testing parameters,
we define the difference $\delta(\mathbf c)$ as the mean difference between the truth parametric solution and the corresponding GAROM prediction, such that:
\begin{equation}
    \delta(\mathbf c) = \frac{1}{N_u}\sum_{i=1}^{N_u}(u_i(\mathbf c) - \hat{u}_i(\mathbf c)),
\end{equation}
where $\mathbf u(\mathbf c) = \begin{bmatrix} u_1(\mathbf c) & \dots & u_{N_u}(\mathbf c)  \end{bmatrix}$ and $\hat{\mathbf u}(\mathbf c) = \begin{bmatrix} \hat{u}_1(\mathbf c) & \dots & \hat{u}_{N_u}(\mathbf c) \end{bmatrix}$.
We highlight that we do not consider the absolute value here in order to detect possible systematic over- and under-estimations of the proposed model.

\begin{figure}[!thb]
    \centering
    \includegraphics[width=\linewidth]{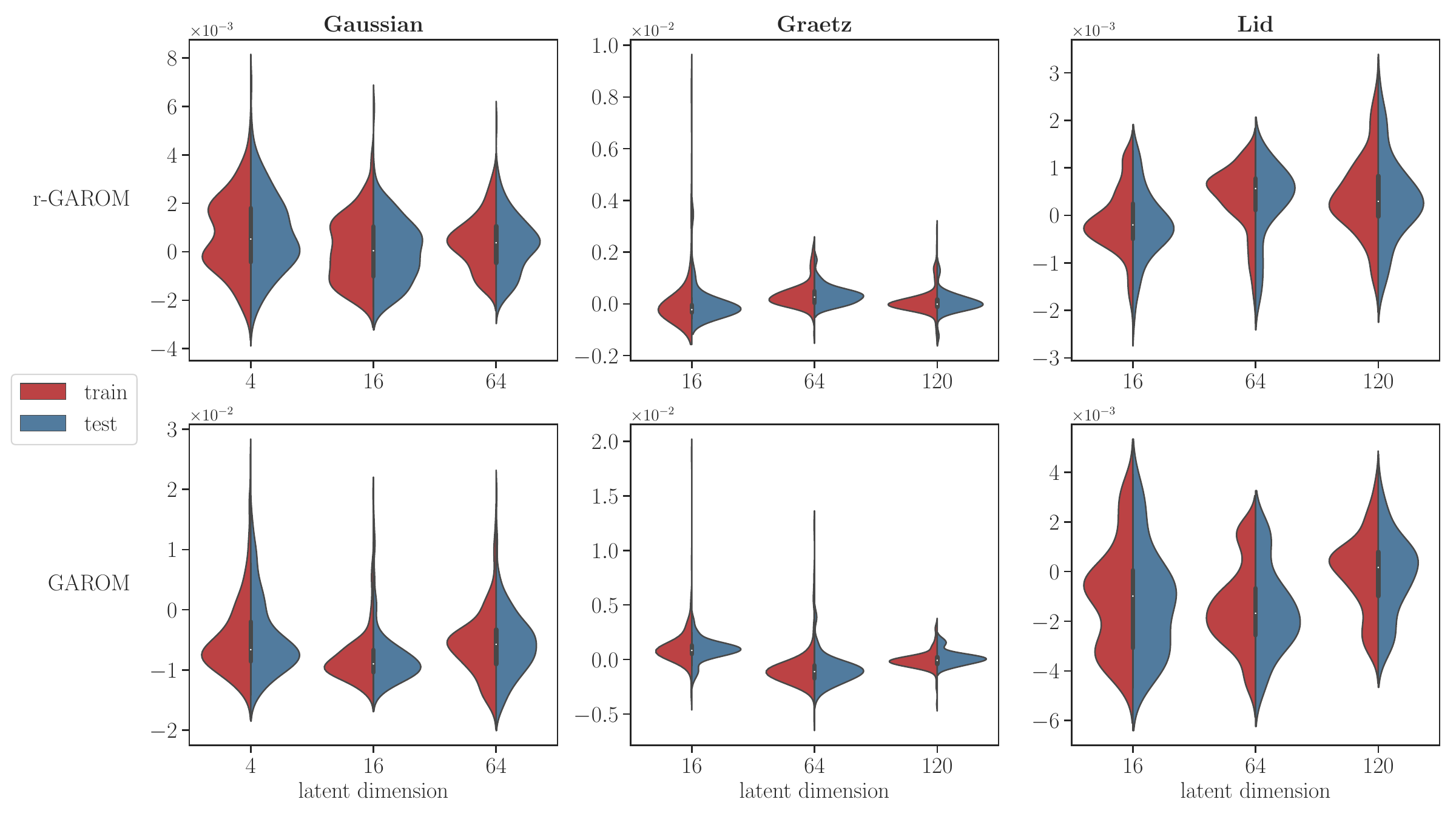}
    \caption{\textbf{Distribution of the $\delta$ difference}. The graph depicts, for each latent dimension, train (red) and test (blue) distribution of the $\delta$. \emph{Top}: r-GAROM model. \emph{Bottom}: GAROM model.}
    \label{fig:violin}
\end{figure}

Figure \ref{fig:violin} shows the distribution of the $\delta$ different for all latent dimensions and for all test cases. 
We can note that the distribution of the error computed over the training parameters is quite similar to the error distribution for the testing parameters in all the numerical investigations.
Such behaviour indicates that both models can correctly generalize outside the training dataset, for the proposed problems. Furthermore, the distributions are centered around zero for both models, indicating a good high-fidelity solution reconstruction, as also shown in the previous Section. Finally, it is worth noticing that r-GAROM distributions are \reviewerB{more close to zero} than \reviewerB{the unregularized counterpart}, as a consequence of the regularizer term addition. 
\begin{figure}[!thb]
    \centering
    \includegraphics[width=\linewidth]{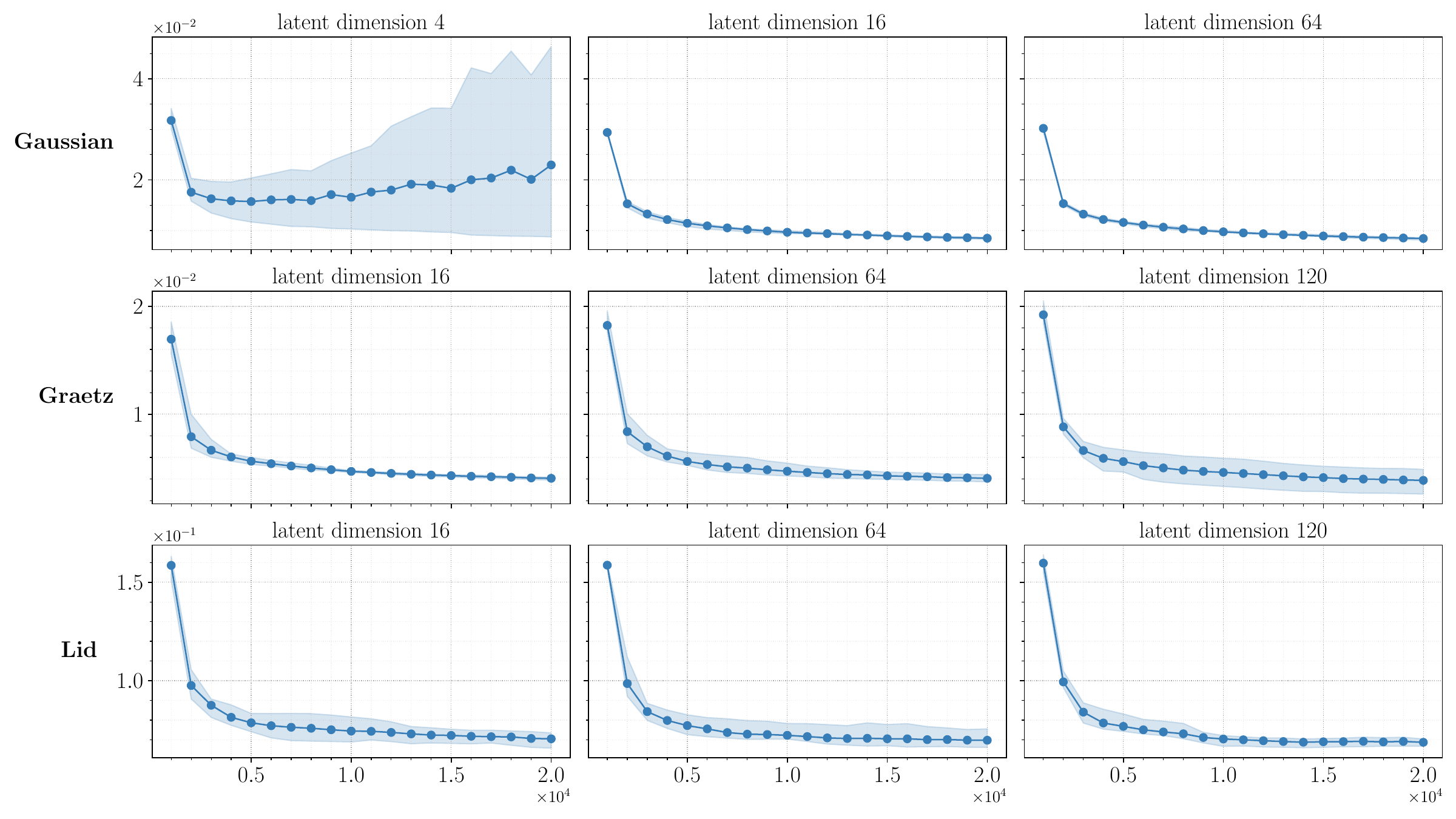}
    \caption{\textbf{r-GAROM convergence graph in $l_2$ relative error for multiple training}. The solid line indicates the average across all simulations. The shaded area represents the interval obtained by taking the maximum and minimum error across all simulations. The total number of training is $5$.}
    \label{fig:convergence}
\end{figure}

The last investigation regards the robustness of the training.
It is well established in the deep learning community that generative adversarial networks tend to be unstable during training~\cite{goodfellow2020generative}. Nevertheless, such behaviour is drastically reduced with BEGAN~\cite{berthelot2017began,conditional_began}. In order to assess the convergence of the optimization procedure, we perform a statistical analysis studying the error trend during the training, using different random initialization for the weights of the network. Practically, $5$ training loops are carried out changing the seed for the random number generator, and using Xavier weight initialization. Figure~\ref{fig:convergence} depicts the mean error and the interval between the minimum and the maximum error across the 5 simulations. The plot indicates a stable convergence of r-GAROM, with sporadic perturbations in the statistical campaign. \reviewerB{The only test case”} that is showing not a stable trend is the Gaussian one, but only when the latent dimension is $4$: we suppose the latent dimension is not sufficiently big to represent the original problem, leading to an aleatory optimization. However, the $l_2$ relative error for the maximum discrepancy is in the same order of magnitude as the average result, showing that the model has still reached a good performance.  

\section{Discussion}
In this study, we proposed GAROM, a generative adversarial \reviewerB{reduced order modeling} approach to tackle parametric PDE problems. The presented methodology is also extended to a regularized version (r-GAROM) in case the solution for the PDE is unique. The methodology is general, and applicable to a variety of ROM problems. \reviewerB{The GAROM and r-GAROM methods are tested} on three parametric benchmarks for different latent dimensions, showing very promising results in data-driven modeling.  Among the deep learning approaches for ROM, r-GAROM outperforms in terms of accuracy in model prediction all of them in seven out of nine tests. Moreover, this approach seems to be more resilient to overfitting. \reviewerB{With respect to POD, it emerges in these examples the POD models have a high difference between train and test errors when the \reviewerB{POD is not able to efficiently capture the underlying behavior represented by the
snapshot} --- e.g. due to nonlinearity.}
Differently, GAROM \reviewerB{and its variation} do not suffer from poor generalization, resulting in a more reliable model, despite the lack of precision in some tests. Surprisingly, it also shows any dependence between the $l_2$ relative mean error and the latent dimension we use in the discriminator, making it less sensitive to this parameter. It must be repeated that, with the GAROM approach, we are not projecting data on a reduced space, but the inner dimension is only used in the discriminator autoencoder.
The adversarial approach combined with the proposed regularisation seems also helpful to stabilize the training in terms of loss trend, making such an approach practically \reviewerB{less difficult to tune
the hyper-parameters}.
Finally, the statistical approach \reviewerB{allows the computation of} the variance learned by the generator, making it possible to quantify the uncertainty of predictions applying different statistical strategies.

We identify multiple exciting research directions for the methodology. Firstly, improving the quality of the neural networks for the generator, discriminator, and conditioning mechanisms, and applying the methodology to more complex problems. As already stated, in the work we have used very simple architectures, which limits the generation capability. As an example, the framework could be easily extended to time-dependent problems, employing LSTM  \cite{hochreiter1997long} or \reviewerB{transformer \cite{vaswani2017attention} networks} for the conditioning mechanism.  Alternatively, as also suggested in \cite{berthelot2017began}, employing different architecture for the discriminator, e.g. U-net \cite{ronneberger2015u}, could lead to accuracy improvement. Another possibility is to enhance the conditioning set with more information, e.g. by passing POD modes, or mesh information. 
Another very interesting \reviewerB{extension} of this work would be the continuous extension of the entire architecture. \reviewerB{With the proposed architecture}, both the networks' dimension scales with the number of components in the snapshots vector. To apply such an approach to very fine meshes, the computational requirement to train and infer the model would be huge and sometimes unfeasible. In these cases, a spatial continuous extension can lead to a wider diffusion, especially in more complex problems. Another possibility to mitigate the difficult application to finer meshes (and so snapshots with more \reviewerB{degrees of freedom}) is the employment of the architecture presented in \cite{coscia2023continuous} where the authors employ continuous convolutional neural networks over unstructured meshes. \reviewerA{Finally, extending to a continuous setting might allow the use of our methodology to learn mappings between functional spaces in a discretization invariant manner, as done with Neural Operators \cite{rahman2022generative, seidman2023variational}.}

In summary, the presented new methodology, combined with the different new research directions, paves the way to the application of GAROM in very complex settings, exhibiting great potential for future ROM developments in computational science.

\newpage
\section{Acknowledgements}
This work is partially supported by European Union Funding for Research and Innovation — Horizon 2020 Program — in the framework of European Research Council Executive Agency: H2020 ERC CoG 2015 AROMA-CFD project 681447 "Advanced Reduced Order Methods with Applications in Computational Fluid Dynamics" P.I. Professor Gianluigi Rozza, by European Union Funding for Research and Innovation --- Horizon Europe Program --- in the framework of European Research Council Executive Agency: ERC POC 2022 ARGOS project 101069319 ``Advanced Reduced order modellinG: Online computational web server for complex parametric Systems'' P.I. Professor Gianluigi Rozza, by European High-Performance Computing Joint Undertaking project Eflows4HPC GA N. 955558, by PRIN "Numerical Analysis for Full and Reduced Order Methods for Partial Differential Equations" (NA-FROM-PDEs) project.

\newpage
\bibliographystyle{abbrv}
\bibliography{biblio}

\end{document}


\maketitle
\thispagestyle{empty} 

\section{Hyperparameter tuning}
\subsection{Hyperparameter tuning baseline models}
We report the model details to reproduce the results presented in Table 1. The optimization is carried out using RayTune, searching by grid-search the number of layers and their sizes across $100$ different models for each neural network and each experimental problem. For POD-NN, AE-NN, AE-RBF, we fixed the latent dimension to four for the Gaussian test, and sixteen for the Graetz and Lid Cavity problem. In this section, $N_u$ indicates the field solution vector dimension and $N_l$ is the latent dimension size.
The activation functions were chosen from a collection of four different activations (ReLU, SiLU, Softplus, Tanh). The Autoencoder is built by choosing the number of layers $L$, and the output dimension of each layer is chosen by dividing (multiplying) each encoding (decoding) layer input dimension by a chosen $F$. In particular we choose  $L \sim U(1, 5)$, and $ F \sim U(2, 5)$. For the Neural Network in POD-NN we built a neural network by choosing the number of layers $L$ and the layer size $S$. In particular, we choose  $L \sim U(1, 8)$, and $ F \sim U(5, 50)$. For DeepONet we built a neural network by choosing the number of layers $L$ and the layer size $S$. In particular, we choose  $L \sim U(1, 4)$, and $ F \sim U(20, 120)$. For NOMAD we use the same hyperparameter optimization as DeepONet since only the aggregation mechanism changes. Finally, no hyperparameter optimization was performed on (r-)cGAN since only the loss are changing with respect to (r-)GAROM, for which we have done hyperparameter optimization (see \ref{sec:hypergarom}).\\

\noindent\textbf{POD-NN}\\
POD is performed by truncated singular value decomposition, with rank equals to the latent dimension size. For the Gaussian test case, the neural network is composed $6$ fully connected layers of size $43$ and ReLU activation function. For the Graetz test case, the neural network is composed of $2$ fully connected layers of size $24$ and $64$ with ReLU activation function. For the Lid Cavity test case, the neural network is composed $6$ fully connected layers of size $23$ and SiLU activation function.\\

\noindent\textbf{AE-NN}\\
For the Gaussian test case, the neural network is composed $5$ fully connected layers of size $45$ and Tanh activation function. The autoencoder comprises five layers of size $[N_u / 2, N_u / 4, N_l, N_u / 4, N_u / 2]$ with Tanh activation.
For the Graetz test case, the neural network is composed of $2$ fully connected layers of size $24$ and $64$ with ReLU activation function. The autoencoder comprises six layers of size $[N_u / 3, N_u / 6, N_l, N_u / 6, N_u / 3]$ with ReLU activation.
For the Lid test case, the neural network is composed $3$ fully connected layers of size $35$ and ReLU activation function. The autoencoder is composed of five layers of size $[N_u / 4, N_u / 16, N_l, N_u / 16, N_u / 4]$ with ReLU activation.\\ 

\noindent\textbf{AE-RBF}\\
The Radial Basis Function kernel is Gaussian with leght scale equals to one. For the Gaussian test case, the autoencoder is composed of two layers of size $N_u / 2$ and $N_u / 4$ with Tanh activation.
The autoencoder is composed of five layers of size $[N_u / 3, N_u / 6, N_l, N_u / 6, N_u / 3$ with ReLU activation.
For the Lid test case, the autoencoder comprises two layers of size $N_u / 4$ and $N_u / 16$ with ReLU activation.\\

\noindent\textbf{DeepONet}\\
The DeepONet consists of two sub-networks, one for encoding the parameters defining the differential problem (branch network), and another for encoding the locations for the output functions (trunk network). We choose the same architectures for the hidden layers of trunk and branch networks. For the Gaussian test case, the networks are composed of two layers of size $115$ with Tanh activation. For the Graetz test case, the networks are composed of three layers of size $104$ with Tanh activation.
For the Lid Cavity test case, the networks are composed of one layer of size $99$ with Tanh activation.\\
                  
\noindent\textbf{NOMAD}\\
The NOMAD network is the same as the DeepONet network with the only difference in how the branch and trunk networks are organised. Specifically, following the same procedure as the one indicated in the original article \cite{seidman2022nomad}, the branch network takes a parameter and outputs its embedding. This is concatenated with the spatial location where the output function is evaluated, and this concatenation is passed to the trunk network.\\

\noindent\textbf{(r-)cGAN}\\
The conditional GAN and its regularized counterpart are composed of a generator and discriminator network. To have a fair comparison with GAROM, we use the same Generator Network as in GAROM (see Table \ref{tab:garom_s}), fixing only the noise dimension to $12$ since we found it beneficial for the training. Equally, the (r-)cGAN discriminator network is the same as the encoder of the Discriminator network (fixing encoding dimension $N_l=120$). Once the encoded variable is concatenated with the conditioning variable, a linear layer is applied, which follows a sigmoid activation to ensure the binary output. Only for the Lid Cavity test the training was done without batching since better performance in terms of training and test error was achieved.

\subsection{Hyperparameter tuning (r-)GAROM}\label{sec:hypergarom}
We report the model details to reproduce the numerics in Table 1 in \emph{Results}. Let the generator activation function be $\sigma_G$, the discriminator activation function be $\sigma_D$ and $N_c^{in}$, $N_c^{in}$, $N_c^{out}$ the hyperparameter to optimize. The general (r-)GAROM architecture is reported in Table \ref{tab:arch}
\begin{table}[h]
    \caption{\textbf{GAROM model architecture}. The table reports the type of layer, its weights and the activation after each layer. The input dimension is indicated as $N_u$, noise dimension $N_z$, conditioning dimension $N_c$, and latent dimension $N_l$.}
    \label{tab:arch}
    \vspace{0.2cm}
\resizebox{\linewidth}{!}{
\begin{tabular}{ccccccccc}
\rowcolor[HTML]{FFFFFF} 
\toprule
{\color[HTML]{000000} \textbf{Layer}} & \multicolumn{2}{c}{\cellcolor[HTML]{FFFFFF}{\color[HTML]{000000} \textbf{Generator}}} & \multicolumn{2}{c}{\cellcolor[HTML]{FFFFFF}{\color[HTML]{000000} \textbf{Discriminator}}} & \multicolumn{2}{c}{\cellcolor[HTML]{FFFFFF}\textbf{\shortstack{Generator \\ Conditioning-net } }}  & \multicolumn{2}{c}{\cellcolor[HTML]{FFFFFF}\textbf{\shortstack{Discriminator \\ Conditioning-net }}}  \\ 
\cmidrule(lr){2-3} \cmidrule(lr){4-5} \cmidrule(lr){6-7} \cmidrule(lr){8-9}
\rowcolor[HTML]{FFFFFF} 
\multicolumn{1}{l}{\cellcolor[HTML]{FFFFFF}{\color[HTML]{000000} }} & {\color[HTML]{000000} \textbf{weights}} & {\color[HTML]{000000} \textbf{activation}} & \multicolumn{1}{c}{\cellcolor[HTML]{FFFFFF}{\color[HTML]{000000} \textbf{weights}}} & \multicolumn{1}{c}{\cellcolor[HTML]{FFFFFF}{\color[HTML]{000000} \textbf{activation}}}  & \multicolumn{1}{c}{\cellcolor[HTML]{FFFFFF}\textbf{weights}} & \multicolumn{1}{c}{\cellcolor[HTML]{FFFFFF}\textbf{activation}}  & \multicolumn{1}{c}{\cellcolor[HTML]{FFFFFF}\textbf{weights}} & \multicolumn{1}{c}{\cellcolor[HTML]{FFFFFF}\textbf{activation}} \\
\midrule
Fully-connected & [$N_z + N_f$, $\sfrac{N_u}{6}$] & $\sigma_G$& [$N_u$, $\sfrac{N_u}{3}$]  & $\sigma_D$& [$N_c$, $N_c^{in}$] & $\sigma_G$ & [$N_c$, $\sfrac{N_l}{2}$] & $\sigma_D$  \\
Fully-connected & [$\sfrac{N_u}{6}$, $\sfrac{N_u}{3}$] & $\sigma_G$ & [$N_c^{in}$, $N_c^{in}$] & $\sigma_D$& [$N_c^{in}$, $N_c^{out}$] & Identity & [$\sfrac{N_l}{2}$, $N_l$] & Identity  \\
Fully-connected & [$\sfrac{N_u}{3}$, $N_u$] & Identity & [$\sfrac{N_u}{6}$, $N_l$] & $\sigma_D$& - & - & - & -   \\

Fully-connected & - & - & [$2N_l$, $\sfrac{N_u}{6}$] & $\sigma_D$& - & - & - & -   \\
Fully-connected & - & - & [$\sfrac{N_u}{6}$, $\sfrac{N_u}{3}$] & $\sigma_D$& - & - & - & -   \\
Fully-connected & - & - & [$\sfrac{N_u}{3}$, $N_u$] & Identity & - & - & - & -   \\

\bottomrule

\end{tabular}}
\end{table}

The optimization is carried out using RayTune, searching by grid-search the number of layers and their sizes across $100$ different models for each neural network and each experimental problem. The activation functions were chosen from a collection of four different activations (ReLU, SiLU, Softplus, Tanh), and the layer sizes as well as the number of layers were chosen accordingly to a random uniform integer distribution. In order to make an ease comparison on convergence and generalization for the test in the section \emph{Generalization}, and \emph{Robustness} we use the same activation functions for discriminator (ReLU) and generator (SiLU) across all tests. Furthermore, only for those two sections, we set $N_c^{out} = 5N_z$, $N_c^{in} = 2 N_z$ , $N_c^{1} = N_z / 4$. For \ref{tab:arch} we report the best architecture we found, so to make a fair comparison with the best architectures for the baseline models.
The details for r-GAROM and GAROM for the different test cases are shown in Table \ref{tab:rgarom_s} and Table \ref{tab:garom_s} respectively.

\begin{table}[h]
\caption{\textbf{r-GAROM hyperparameters}. The table reports the r-GAROM hyperparameters used in the tests in \emph{Results}.}
    \centering
    \begin{tabular}{cccc}
    \toprule
         & \textbf{Gaussian} & \textbf{Graetz} & \textbf{Lid Cavity} \\
    \midrule
     $\sigma_G$ & SiLU & ReLU & ReLU\\
     $\sigma_D$ & Tanh & Tanh & ReLU\\
     $N_c^{in}$ & 17 & 31 & 29\\ 
     $N_c^{out}$ & 35 & 63 & 58\\
     \bottomrule
    \end{tabular}
    \label{tab:rgarom_s}
\end{table}
\begin{table}[h]
\caption{\textbf{GAROM hyperparameters}. The table reports the GAROM hyperparameters used in the tests in \emph{Results}.}
    \centering
    \begin{tabular}{cccc}
    \toprule
         & \textbf{Gaussian} & \textbf{Graetz} & \textbf{Lid Cavity} \\
    \midrule
     $\sigma_G$ & SiLU & SiLU & SiLU\\
     $\sigma_D$ & Tanh & ReLU & ReLU\\
     $N_c^{in}$ & 17 & 24 & 24\\ 
     $N_c^{out}$ & 35 & 60 & 60\\
     \bottomrule
    \end{tabular}
    \label{tab:garom_s}
\end{table}

\section{Baseline Models training details}
In AE-NN, AE-RBF the autoencoder is trained for $1000$ epochs, while in AE-NN and POD-NN the artificial neural network is trained for $20000$ epochs. DeepONet and NOMAD baselines are both trained for $1000$ epochs. The (r-)cGAN model is trained for $20000$ epochs for the Gaussian and Graetz test case, while $2500$ epochs were enough to ensure a stable convergence for the Lid Cavity experiment.  All the baseline neural networks-based models are optimized with Adam \cite{kingma2014adam} using $0.001$ learning rate, minimizing the mean square error loss, and training until stable convergence is evidenced using a batch size of $4$, otherwise not stated.

\bibliography{biblio}